\documentclass[runningheads,a4paper]{llncs}

\usepackage{amssymb,amsmath}
\usepackage{graphicx}

\usepackage{url}
\urldef{\mailsa}\path|{lpalma, lrargerich}@fi.uba.ar|    
\newcommand{\keywords}[1]{\par\addvspace\baselineskip
	\noindent\keywordname\enspace\ignorespaces#1}

\begin{document}
	
	\mainmatter  % start of an individual contribution
	
	% first the title is needed
	\title{Iterative evaluation of LSTM cells}
	
	% a short form should be given in case it is too long for the running head
	\titlerunning{Iterative evaluation of LSTM cells}
	
	% the name(s) of the author(s) follow(s) next
	%
	% NB: Chinese authors should write their first names(s) in front of
	% their surnames. This ensures that the names appear correctly in
	% the running heads and the author index.
	%
	\author{Leandro Palma%
		\and Luis Argerich}
	%
	% (feature abused for this document to repeat the title also on left hand pages)
	
	% the affiliations are given next; don't give your e-mail address
	% unless you accept that it will be published
	\institute{Departamento de Computacion, Facultad de Ingenieria, Universidad de Buenos Aires,\\
		Paseo Colon 850, Ciudad Autonoma de Buenos Aires, Argentina,
		Universidad Nacional de Tres de Febrero (UNTREF)\\
		\mailsa\\
		\url{http://www.fi.uba.ar}}
	
	%
	% NB: a more complex sample for affiliations and the mapping to the
	% corresponding authors can be found in the file "llncs.dem"
	% (search for the string "\mainmatter" where a contribution starts).
	% "llncs.dem" accompanies the document class "llncs.cls".
	%
	
	\maketitle
	\begin{abstract}
		In this work we present a modification in the conventional flow of information through a LSTM network, which we consider well suited for RNNs in general. The modification leads to a iterative scheme where the computations performed by the LSTM cell are repeated over a constant input and cell state values, while updating the hidden state a finite number of times.
		We provide theoretical and empirical evidence to support the augmented capabilities of the iterative scheme and show examples related to language modeling.\\
		The modification yields an enhancement in the model performance comparable with the original model augmented more than 3 times in terms of the total amount of parameters.
		\keywords{Dynamic Systems, RNNs, LSTM cells, Language Modeling.}
	\end{abstract}
	\section{Introduction}
	The study of deep neural network architectures evidenced their efficiency to approximate certain families of complex functions of the input domain\cite{delalleau2011shallow,pascanu2013number}. This characteristic is contrasted with the difficulty of training these neworks using the standard back-propagation algorithm\cite{he2016deep}. We present a model that aims to maintain the flexibility of the deep neural networks without producing these problems. Our model increases the expressive power of a recurrent neural network in a way that is comparable with a deep model with the simplicity of training a shallow model.
	
	Our work aims to introduce a modification in the structure of the LSTM cell block, theoretically-driven by the study of subsequent evaluations of repeating the evaluation of the LSTM cell, updating the hidden state while keeping the input and cell state constant. \\
	We argue that this iterative process defines a dynamic system that accounts for the evolution of the hidden state of the network, driving it to compact regions where the information is presumably retained better. The theory behind the dynamic system studied will be explained in the next sections.
	
	The principal motivation for the proposed modification is the theoretical evidence that the dynamic systems could retain information in the form of an analog state vector. This vector is confined to the basin of attraction corresponding to one of the possible states that are relevant to the resolution of the task\cite{bengio1994learning}. On the other hand, nonlinear dynamic systems, as the one defined by our model, are capable of defining a complex behavior over the phase plane that is useful to flatten the hidden class manifolds.
	
	To decide the number of iterations to perform we found that the usage of a simple logistic regression over the LSTM cell block variables, thresholded to select whether or not to modify the current hidden state by an additional iteration, could serve as a controller that is properly optimized to select whether or not to perform an additional iteration. This simple model avoid the need to select the number of iterations as an hyper-parameter of the model.\\
	This controller is inspired by the gating mechanisms already present in LSTM cells and its goal is to weight the importance of the iterations in the final result.\\
	We provide an open-source implementation\footnote{\texttt{https://github.com/PalmaLeandro/iterativeLSTM}} of the LSTM cell network modified as is proposed, along with empirical evidence of the improvements in the performance of the modified models compared to its former baseline on the task of language modeling.
	\section{Related work}
	One clear example of an architecture of NN that introduces the iterative scheme is LoopyNN presented in \cite{caswellloopy}. This recurrent model performs subsequent iterations using the last result as input for the next iteration.\\
	While this behavior is similar to our proposal, differences can be drawn on the method and on the theory behind it.
	Nevertheless, much of the characteristics of our model are noted by Caswell and his colleagues. We note that the amount of unrolls of its architecture is related to the amount of iterations performed by our model.
	
	An iterative scheme similar to the proposed one is studied by \cite{liao2016bridging}. That work covers the improvements in the performance of a fully connected NN that performs several evaluations after every new information is exposed to the network. Moreover, such work relates the iterative scheme with recurrent structures found in the human brain.
	The concept of \textit{readout time} presented in that work is related to the number of iterations performed by our model. The relation among these concepts is supported by the similar enhancements found in model's performance while increasing their values. 
	
	The analysis performed by \cite{laurent2016recurrent} explores the chaotic properties of LSTM cells as dynamic systems. Even when the conditions over which its LSTM networks are evaluated differ from real applications such work motivated the study of the non-linear components of our model in order to avoid the chaotic behavior exhibited and which would prevent the network state to converge towards stable configurations.
	\section{Iterative LSTM cell}
	We considered how to fold models that share the weights and state of several layers of LSTM cells to yield a more compact representation that executes a fixed number evaluations.
	
	In order to optimize the model parameters using gradient based methods the state exposed by the network has to be a real-valued vector. Thus, the latching of information in the system is accomplished by the evolution of such vector towards stable configurations\cite{bengio1994learning}. This evolution is governed by the dynamics of a non-autonomous system which is defined by the model's formulation and is parameterized by the input values at every time-step\cite{pascanu2013difficulty}.
	
	For the iterative scheme proposed, the evolution of the network state at every time-step is governed by an autonomous system which would last as long as the model performs additional iterations.\\
	The complete evolution of the network state is the aggregation of the several state changes achieved at every time-step.  
	Such representations are easier to classify by adjacent components due to the sharp frontiers with the set of states that belongs to other attraction basins, and therefore different information.\\
	We note that a model whose state evolution is subject to the conditions mentioned has to organize its fixed points to flatten manifolds into the proper configurations as well as connect or maintain these domains across subsequent time-steps in order to unambiguously recall information.
	
	One important observation about dynamic systems derived from the existence and uniqueness theorem is that the trajectories cannot intersect each other and, over convergent conditions, this shall produce the contraction of the volume of states being attracted. Hence, the contracted volume of states exhibits a simpler surface.
	This behavior yields sparse and compact regions as more flattened manifolds to be classified by higher layers. 
	Under the stated conditions we expect an increase in the performance of the model modified as is proposed, with respect to its original version. This is supported by the fact that the diffuse or nonlinear edges of latent manifolds jeopardize the model's performance because of the limitations exhibited by recurrent neural models\cite{elman1990finding}.
	
	In order to achieve the desired convergent conditions we propose a modification to the LSTM structure. This modification aims to induce an autonomous dynamic system whose state, given by the hidden state $h$ of the LSTM block, converges to a confined domain closer to the attractor of the basin in which the initial conditions reside. These initial conditions are given by the value of the hidden state at previous time-step.
	
	The model resulting from applying the proposed modification to a LSTM network could be summarized by the following ecuations calculated in the given order, for every iteration $\tau$.\\
	\begin{equation*}
		\begin{split}
			j(\tau) &= tanh(W_{rec, j} h +W_{in, j} x(t) + b_{j} ) =  tanh(W_{rec, j} h + C_j)\\
			i(\tau) &= \sigma(W_{rec, i} h + W_{in, j} x(t) + b_{j} ) =  \sigma(W_{rec, i} h + C_i)\\
			f(\tau) &= \sigma(W_{rec, f} h + W_{in, j} x(t) + b_{j} ) =  \sigma(W_{rec, f} h + C_f)\\
			c(\tau) &= f(\tau)c(t-1) + i(\tau)j(\tau) = f(\tau) c_0 + i(\tau)j(\tau) \\
			o(\tau)& = \sigma(W_{rec, o} h + W_{in, j} x(t) + b_{j} ) = \sigma(W_{rec, o} h + C_o)
		\end{split}
	\end{equation*}
	\begin{equation}\label{iterative_LSTM_cell_formulation}
		\begin{split}
			h(\tau) &= o(\tau) \, tanh(c(\tau))\\
			y(t) &= h(\tau) + x(t)
		\end{split}
	\end{equation}
	As a result, the evolution of the state $h$ through the iterations performed describes a dynamic system. Its phase plane at any time-step $t$ of the input sequence is defined by the former internal cell state $c_0$ and the input values $x(t)$ since the constants $C_\varphi \, , \, \forall \,  \varphi \, \in \, \{i, j, f, o\}$ depend on such information.
	
	An iteration activation gate $p$ is introduced with the idea of selecting whether to perform an additional iteration and modify the hidden state or expose it to the next layers. This gate consists of a logistic regression of the inputs, the recently calculated state $h$ and the internal gate variables $i$, $j$ and $f$. Its output is then compared with a threshold which is a parameter of the iteration and varies to more restrictive values to reduce the overall iterations made without constraining the cell to an arbitrary limit.
	
	Another modification introduced in our model is the resolution of a residual mapping of the inputs\cite{he2016deep}. This customization of the model is supported by the results found in \cite{he2016deep} where the addition of a direct connection of the input allows to reference the result of a deep neural network inference with respect to its input.\\
	Moreover, \cite{liao2016bridging} stated the similarity of the networks that share weights across the depth dimension with residual networks.\\
	Additionally, in \cite{caswellloopy} the residual mapping is tested experimentally and evidence is found to support its application on neural networks with shared parameters.
	\section{Convergence of Iterative LSTM cells}
	The formulation of RNNs produces a well-known dynamic system studied by \cite{pascanu2013difficulty} and others \cite{bengio1994learning}. Is in the publication of Pascanu et. al. that the effect of the non-autonomous components of the system is explored to expose the challenges that it represents to gradient based methods. Its effect has to be considered in order to produce the intended convergence of the state to the corresponding attractors.\\
	In the proposed iterative scheme this is achieved by maintaining a constant input and cell state values through all the iterations performed by the model within a time step of the input sequence. This constraint yields an autonomous dynamic system whose dynamics are determined by the inputs and cell state values, at a particular time-step $t$ of the input sequence.
	
	Then, considering subsequent evaluations of an LSTM network over a constant input $x$ and cell state $c$ values while varying its hidden state $h(\tau)$ at discrete steps $\tau$ yields the folowing dynamic system
	\begin{equation}\label{iterative_LSTM_cell_dynamic_formulation}
			\frac{\partial h(\tau)}{\partial \tau} = h(\tau + 1) - h(\tau) = LSTM(x(t), c(t), h(\tau)) - h(\tau) = g_t(h(\tau)) - h(\tau)
	\end{equation}
	where the $LSTM$ function corresponds to the calculations required to update the hidden state of the vanilla LSTM network.\\
	This system has been studied by \cite{laurent2016recurrent}, exposing its behavior for more than 200 iterations over a null value of the inputs. The results extracted from that analysis were that the sensibility of the system to variations in the initial conditions produce chaotic trajectories of states. \\
	This implies that sightly different initial states could produce a completely different final state by projecting the behavior of the model long enough. Consequently, the predictions of the model for similar inputs may as well be different.
	
	The following result is intended to provide sufficient conditions to bound the system's Liapunov coefficients to a subset that leads to a coherent behavior of the model for variations in the initial conditions\cite{StrogatzBook}. By meeting these conditions the difference on the predictions made for slightly different initial states shall be bounded.
	\begin{theorem}
		\label{iterative_LSTM_cell_convergence_theorem}
		Let $f(x(t), h(t-1), c(t-1))$ be a model consisting of a LSTM cell block such that
		\begin{equation}
			\begin{split}
				h(t) &= f(x(t), c(t-1), h(t-1)) =  LSTM(x(t), c(t-1), h(t-1))
			\end{split}
		\end{equation}
		where $x(t)$ is the value of the input sequence at the time-step $t$. $c(t-1)$ and $h(t-1)$ are vectors with the values of the internal and exposed state at the previous time-step, respectively.\\
		The subsequent evaluations of $f$ implicitly defines the dynamic system
		\begin{equation}\label{iterative_LSTM_cell_convergence_theorem_dynamic_system}
			\begin{split}
				\dfrac{\partial h(\tau)}{\partial \tau} &= h(\tau + 1) -  h(\tau) = f(x(t), c(t-1), h(\tau)) - h(\tau) = g_t(h(\tau)) - h(\tau)
			\end{split}
		\end{equation}
		where $g_t(h(\tau))$ is a function analog to $f(x(t), c(t-1), h(t-1))$ where $x(t)$ and $c(t-1)$ are kept constant. Under these conditions the following implication holds
		\begin{equation}
			\sigma_j + \frac{1}{4} \sigma_i + \frac{1}{4} \sigma_f + \frac{1}{4} \sigma_o < 1 \implies \lambda_i < 0 \, , \, \forall \, i \in 0,...,n-1
		\end{equation}
		where $\sigma_j$, $\sigma_i$, $\sigma_f$ and $\sigma_o$ are the principal singular values of the matrices that weights the recurrent connections of the $j$, $i$, $f$ and $o$ gates of the LSTM cell block, respectively. $\lambda_i$ is the Liapunov exponent of the dynamic system defined by (\ref{iterative_LSTM_cell_convergence_theorem_dynamic_system}) at the $i$-th dimension. $n$ is the number of cell units.
	\end{theorem}
	\begin{proof}
		See appendix.
	\end{proof}
	The principal implication of the theorem \ref{iterative_LSTM_cell_convergence_theorem} is that the evolution of the hidden state over the iterative scheme proposed is not chaotic for a particular set of the model's parameters.

	An initial configuration of network parameters that matches the conditions of theorem \ref{iterative_LSTM_cell_convergence_theorem} could be achieved following the suggestions for initialization in \cite{zilly2016recurrent} derived from the Geršgorin circle theorem. Moreover, as mentioned in the publication of Zilly et. al., the L1 and L2 regularization techniques could be used to enforce the conditions required for the application of the theorem \ref{iterative_LSTM_cell_convergence_theorem}.
	
	We believe that the presented analysis could be extended to RNNs in general providing more evidence that the iterative scheme proposed enhances the capabilities of other recurrent models as well.
	\section{Experiments}
	Following the configurations in \cite{RegularizationZaremba} several experiments were performed training different architectures of RNNs as language models. 
	We used the implementation of the publication of Zaremba et. al. that was released with the TensorFlow library\footnote{Code available at\\
		\parbox[t]{10em}{https://github.com/tensorflow/models/tree/master/tutorials/rnn/ptb}}, as the base for our experiments.\\
	The corpus used to train the model is the Penn Treebank \cite{PennTreebankMarcus}\footnote{Corpus available at \texttt{http://www.fit.vutbr.cz/~imikolov/rnnlm/simple-examples.tgz}.}. \\
	The base arquitecture used for the recurrent network is the corresponding with the `medium' size model presented in \cite{RegularizationZaremba}.
	The medium size model consist of the embedding projection layer, 2 layers of LSTM blocks containing 650 units each and a final softmax layer. The large model architecture is the same but each layer of LSTM cells contained 1500 units and its performance is reported as in its original publication.\\
	The embedding projection layer, the layers of the RNN and the softmax layer are connected through regularization connections that applied the Dropout\cite{srivastava2014dropout} method with a probability of keeping the connection of $0.5$. 
	The parameters were optimized using minibatch gradient descent with a batch size of $20$. These parameters were initialized by a random uniform distribution at the interval $[-0.5, 0.5]$.
	The gradients calculated to reduce the loss function are clipped to a norm value of $5$.
	The training regime consisted of 39 epochs where for the first 6 epochs a learning rate of $1.0$ was set and then this value was reduced by a factor of 1.2 for the remaining epochs.
	
\begin{figure}
		\centering
		\includegraphics[scale=0.2150]{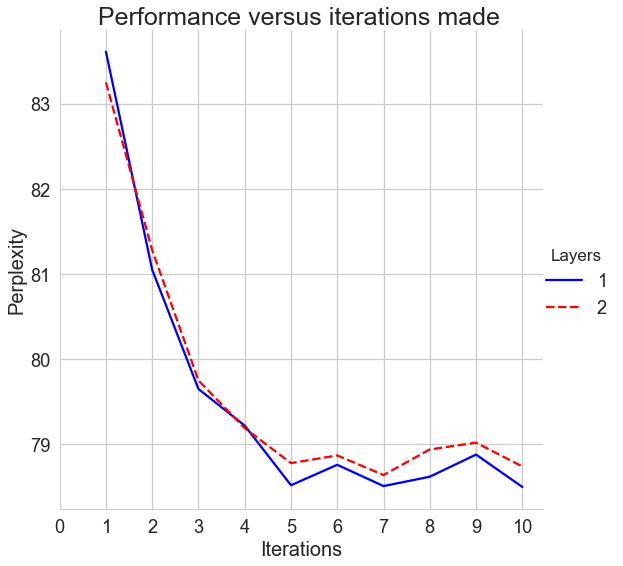}
		\caption{Perplexity per word as function of the iterations executed at every time-step.}
		\label{ppl_vs_iterations_figure}
	\end{figure}

	Figure \ref{ppl_vs_iterations_figure} presents the results of the experiments performed fixing the amount of iterations executed to incremental values.
	The results of the experiments shows that there is an enhancement in the proposed model's performance as the number of iterations performed increases.
	
	As is visible in the Figure \ref{ppl_vs_iterations_figure} the reduced model constituted by one layer of iterative LSTM cells consistently outperforms its augmented version as the number of iterations increase. We attribute this difference in the performance of the models to the overfitting suffered by the augmented model whose expressive capabilities were improved over the iterative regime and the inclusion of an additional layer.\\
	Such observation supports the conjecture that the iterative scheme proposed improves the expressiveness of the model.
	
	Note that in the case of the execution of a single iteration, where the modified network is comparable to its original form, the performance of the smaller model is worse than the augmented model and this effect is inverted as more iterations are executed. This indicates that the improvements over the models' performance are not caused by the modifications made to its structure, namely the iteration activation gate and the residual mapping chosen, rather than by the application of the iterative scheme proposed.
	
	Table 1 presents the performance of the trained models in terms of perplexity per word, where lower is better.
	\begin{center}
		\label{models_experiment_performance_table}
		Table 1: Perplexity per word on Penn Treebank corpus.
		\begin{tabular}[t]{| c  | c | c |}
			\hline
			Model & Size & Perplexity on Test Set\\
			\hline
			\hline
			LSTM & 16M  & 84.48\\
			\hline
			Iterative LSTM & 16M  & 78.46\\
			\hline
			\hline
			LSTM(2 layers) & 20M  & 83.25\\
			\hline
			Iterative LSTM(2 layers) & 20M  & 78.60\\
			\hline
			\hline
			LSTM(2 layers)& 50M & \textbf{78.29}\\
			\hline
		\end{tabular}
	\end{center}
	The results reported in table 1 expose that the performance achieved by the model that applies the proposed modification is comparable with larger versions with more than 3 times the total amount of parameters of our model.
	\section{Conclusions}
	In this work we studied a modification over the traditional LSTM structure that produces an iterative scheme where the inference is done incrementally. 
	We presented theoretical evidence to support the proposed scheme based on the study of the dynamic system defined by the iterative evaluation of the recurrent network.\\
	The results of the experiments executed to expose the effect of the proposed modification supports the theoretical motivation that lead to the development of the presented model.\\
	A comparison of the performance achieved by our model showed a capacity comparable to its largest original version, augmented up to 3 times in terms of the total amount of parameters.
\section{Appendix}
\subsection{Proof of theorem 1} \label{iterative_LSTM_cell_convergence}
\begin{proof}
	The execution of subsequent evaluations of the proposed model yields the dynamic system defined by (\ref{iterative_LSTM_cell_dynamic_formulation})
	\begin{equation*}
			\frac{\partial h(\tau)}{\partial \tau} = h(\tau + 1) - h(\tau) = LSTM(x(t), c(t), h(\tau)) - h(\tau) = g_t(h(\tau)) - h(\tau)
	\end{equation*}
	Then, considering an infinitesimal small variation on the initial conditions $\delta_0$ we look for the variations $\delta_\tau$ on the result of the system after $\tau$ evaluations which would be
	\begin{equation*}
			|\delta_\tau| = |\delta_0| e ^{\tau\lambda} \iff \frac{|\delta_\tau|}{|\delta_0|} =  e ^{\tau\lambda} \iff \lambda = \dfrac{1}{\tau} \ln\left\lvert\dfrac{\delta_\tau}{\delta_0}\right\rvert
	\end{equation*}
	Where $\lambda$ is a vector such that it's coordinates $\lambda_i$ are the Liapunov exponents of the dynamic system at the $i$-th dimension of the vector state $h(\tau)$.
	\begin{equation*}
		\begin{split}
			\lambda = \dfrac{1}{\tau} \ln\left\lvert\dfrac{\delta_\tau}{\delta_0}\right\rvert \iff	\lambda = \dfrac{1}{\tau} \ln \left\lvert \dfrac{g^\tau(h(t-1) + \delta_0) - g^\tau(h(t-1))}{\delta_0} \right\rvert \iff \lambda = \dfrac{1}{\tau} \ln | (g^\tau)' |
		\end{split}
	\end{equation*}
	$g^\tau(h)$ is the application of the $g$ function, defined at (\ref{iterative_LSTM_cell_dynamic_formulation}), $\tau$ times over the state vector $h$.\\
	Then, requiring the Liapunov coefficients to have values that imply the convergence of the sequence of state values yields
	\begin{equation*}
		\dfrac{1}{\tau} \ln | (g^\tau)' | < 0 \iff | (g^\tau)' | < 1 \iff \lambda_i < 0  \, , \, \forall \, i \in 0,...,n-1\\
	\end{equation*}
	Where $\lambda_i$ is the Liapunov exponent of the dynamic system defined by (\ref{iterative_LSTM_cell_dynamic_formulation}) at the $i$-th dimension. $n$ is the number of cell units.
	Next, it is possible to derive $(g^\tau)'$ applying the chain rule to obtain
	\begin{equation*}
		(g^\tau)' = \prod_{i = 0}^{\tau}g'(h(i))
	\end{equation*}
	Replacing this identity on the conditions required for the convergence of the dynamic system in every dimension yields
	\begin{equation*}
		\begin{split}
			| (g^\tau)' | < 1 \iff \left\lvert  \prod_{i = 0}^{\tau}g'(h(i)) \right\rvert < 1 &\iff |g'(h(i))|  < 1 \quad i \in 0, ... , \tau\\
																													     		   &\iff \lambda_i < 0  \, , \, \forall \, i \in 0,...,n-1
		\end{split}
	\end{equation*}
	
	Meanwhile, resolving the derivative of the variables $j, i, f, o$ and $g$, defined at (\ref{iterative_LSTM_cell_formulation}), with respect to a generic state vector $h$ yields
	\begin{equation*}
		\begin{split}
			\dfrac{\partial j(h)}{\partial h} = & \, (1- tanh^2(W_{rec,j} \, h + C_j))W_{rec,j}\\
			\dfrac{\partial i(h)}{\partial h} = & \, \sigma(W_{rec,i} \, h + C_i) (1-\sigma(W_{rec,i} \, h + C_i)) W_{rec,i}\\
			\dfrac{\partial f(h)}{\partial h} = & \, \sigma(W_{rec,f} \, h + C_f) (1-\sigma(W_{rec,f} \, h + C_f)) W_{rec,f}\\
			\dfrac{\partial o(h)}{\partial h} = & \, \sigma(W_{rec,o} \, h + C_o) (1-\sigma(W_{rec,o} \, h + C_o)) W_{rec,o}\\
			\dfrac{\partial g(h)}{\partial h} = & \, \dfrac{\partial o(h)}{\partial h} \, tanh( \, c(h) \, ) + o(h) \, (1 - tanh( \, c(h) \, )^2) \, \dfrac{\partial f(h)}{\partial h} \, c_0\\
			&+ o(h) \, (1 - tanh( \, c(h) \, )^2) \, \dfrac{\partial i(h)}{\partial h} \, j(h)+ o(h) \, (1 - tanh( \, c(h) \, )^2) \, \dfrac{\partial j(h)}{\partial h} \, i(h)
		\end{split}
	\end{equation*}
	where $\frac{\partial tanh(x)}{\partial x} = 1 - tanh(x)^2$ and $\frac{\partial \sigma(x)}{\partial x} = \sigma(x) (1-\sigma(x))$ are diagonal matrices whose coefficients $\frac{\partial tanh(x)}{\partial x}_{i,i}$ and $\frac{\partial tanh(x)}{\partial x}_{i,i}$ corresponds to the evaluation of the functions $tanh'(x_i)$ and $\sigma'(x_i)$ over the $i$-th coordinate of the vector $x$, respectively. Therefore, replacing the definition of $g'(h)$ obtained from the conditions that Liapunov coefficients have to hold in order to produce the intended convergence yields the inequity
	\begin{equation*}
		\begin{split}
			 \lvert \dfrac{\partial o(h)}{\partial h} \, tanh( c(h) \, ) + o(h) \, (1 - tanh( \, c(h) \, )^2) \, \dfrac{\partial f(h)}{\partial h} \, c_0+\quad\quad&\\
			 o(h) \, (1 - tanh( \, c(h) \, )^2) \, \dfrac{\partial j(h)}{\partial h} \, i(h) \rvert < 1 \iff &\lambda_i < 0\\
			\left\lvert \dfrac{\partial o(h)}{\partial h} \right\rvert \, \underbrace{| tanh( c(h) \, )|}_{\le1} + \underbrace{| o(h) |}_{< 1}\,\underbrace{| (1 - tanh( \, c(h) \, )^2) |}_{<1} \, \left\lvert \dfrac{\partial f(h)}{\partial h} \right\rvert \, \underbrace{| c_0|}_{<1} +&\underbrace{| o(h) |}_{< 1}\,\underbrace{| (1 - tanh( \, c(h) \, )^2) |}_{<1} \\
			\left\lvert \dfrac{\partial i(h)}{\partial h} \right\rvert \, \underbrace{| j(h)|}_{<1} + \underbrace{| o(h) |}_{< 1}\,\underbrace{| (1 - tanh( \, c(h) \, )^2) |}_{<1} \, \left\lvert \dfrac{\partial j(h)}{\partial h} \right\rvert \, \underbrace{| i(h)| }_{<1}< 1 &\implies \lambda_i < 0\\
			\left\lvert \dfrac{\partial o(h)}{\partial h} \right\rvert + \left\lvert \dfrac{\partial f(h)}{\partial h} \right\rvert + \left\lvert \dfrac{\partial i(h)}{\partial h} \right\rvert + \left\lvert \dfrac{\partial j(h)}{\partial h} \right\rvert <1 \implies \lambda_i < 0  &\, , \, \forall \, i \in 0,...,n-1
		\end{split}
	\end{equation*}
	Finally, replacing the derivatives of the variables $j, i, f, o$ and $g$ with respect to the vector state $h$ yields
	\begin{equation*}
		\begin{split}
			&\underbrace{|\sigma(W_{rec,0} \, h + C_i) (1-\sigma(W_{rec,o} \, h + C_o)) |}_{\le \frac{1}{4}} \, |W_{rec,o}|+ \underbrace{|\sigma(W_{rec,f} \, h + C_i) (1-\sigma(W_{rec,f} \, h + C_f)) |}_{\le \frac{1}{4}} \, |W_{rec,f}|+\\
			&\underbrace{|(1- tanh^2(W_{rec,j} \, h + C_j))|}_{\le 1} |W_{rec,j}| + \underbrace{|\sigma(W_{rec,i} \, h + C_i) (1-\sigma(W_{rec,i} \, h + C_i)) |}_{\le \frac{1}{4}} \, |W_{rec,i}| <1\\
			&\implies \frac{1}{4} \, |W_{rec,o}| + \frac{1}{4} \, |W_{rec,f}| + |W_{rec,j}| + \frac{1}{4} \, |W_{rec,i}| <1 \implies \lambda_i < 0  \, , \, \forall \, i \in 0,...,n-1
		\end{split}
	\end{equation*}
	Hence, if the following condition holds for the matrices $W_{rec,j}, W_{rec,i}, W_{rec,f}, W_{rec,o}$ the dynamic system, produced by subsequent evaluations over the same input and cell state values while varying the hidden state, produces convergent results for similar initial conditions
	\begin{equation*}
			\sigma_j + \frac{1}{4} \sigma_i + \frac{1}{4} \sigma_f + \frac{1}{4} \sigma_o <1 \implies \lambda_i < 0  \, , \, \forall \, i \in 0,...,n-1
	\end{equation*}
	where $\sigma_j , \sigma_i , \sigma_f , \sigma_o $ are principal singular values of the matrices $W_{rec,j}, W_{rec,i}, W_{rec,f}, W_{rec,o}$ respectively.
\end{proof}
\bibliographystyle{splncs03}
\bibliography{example_paper}
\end{document}